\documentclass[11pt,a4paper]{article}
\usepackage{naaclhlt2018}

\newif\ifcomment\commentfalse

\usepackage{mdwlist}
\usepackage{latexsym}
\usepackage{nicefrac}
\usepackage{booktabs}
\usepackage{amsfonts}
\usepackage{bold-extra}
\usepackage{amsmath}
\usepackage{dsfont}
\usepackage{amssymb}
\usepackage{bm}
\usepackage{graphicx}
\usepackage{mathtools}
\usepackage{microtype}
\usepackage{multirow}
\usepackage{multicol}
\usepackage{latexsym,comment}

\newcommand{\gem}[1]{\mbox{\textsc{gem}}}

\newcommand{\hidetext}[1]{}
\newcommand{\ignore}[1]{}

\ifcomment
\newcommand{\nahocomment}[1]{
\colorbox{yellow}{   \parbox{.8\linewidth}{\scriptsize NAHO: #1}  }}
\newcommand{\shcomment}[1]{ \colorbox{green}{  \parbox{.8\linewidth}{ SH:  #1}}}
\newcommand{\mjpcomment}[1]{   \colorbox{blue}{MJP: #1}}
\newcommand{\jbgcomment}[1]{  \colorbox{red}{   \parbox{.8\linewidth}{ JBG: #1}  }}
\newcommand{\ffcomment}[1]{  \colorbox{yellow}{   \parbox{.8\linewidth}{ FF: #1}  }}
\newcommand{\fpcomment}[1]{  \colorbox{green}{   \parbox{.8\linewidth}{ FP: #1}  }}

\newcommand{\yhcomment}[1]{  \colorbox{green}{  \parbox{.8\linewidth}{ YH:  #1}
}}
\newcommand{\hhecomment}[1]{  \colorbox{blue}{  \parbox{.8\linewidth}{ HH:  #1}
}}
\newcommand{\jjmcomment}[1]{  \colorbox{green}{  \parbox{.8\linewidth}{ John:  #1}
}}
\newcommand{\tncomment}[1]{  \colorbox{blue}{  \parbox{.8\linewidth}{ TN:  #1}
}}
\newcommand{\mnicomment}[1]{  \colorbox{green}{  \parbox{.8\linewidth}{ Mohit:  #1}  }}
\newcommand{\prcomment}[1]{  \colorbox{lightblue}{  \parbox{.8\linewidth}{ Pedro:  #1}  }}

\newcommand{\jszcomment}[1]{  \colorbox{green}{  \parbox{.8\linewidth}{ JSG:  #1}  }}
\newcommand{\ascomment}[1]{  \colorbox{blue}{  \parbox{.8\linewidth}{ AS:  #1}
}}
\newcommand{\vecomment}[1]{  \colorbox{blue}{  \parbox{.8\linewidth}{ VE:  #1}  }}
\newcommand{\halcomment}[1]{  \colorbox{magenta!20}{  \parbox{.8\linewidth}{ Hal:  #1}  }}
\newcommand{\kgcomment}[1]{  \colorbox{blue}{  \parbox{.8\linewidth}{ Kim:  #1}  }}
\newcommand{\vancomment}[1]{
\colorbox{green}{  \parbox{.8\linewidth}{ VAN:  #1}  }}

\newcommand{\alvincomment}[1]{  \colorbox{cyan}{  \parbox{.8\linewidth}{ Alvin:  #1}  }}
\newcommand{\reviewercomment}[1]{  \colorbox{blue}{  \parbox{.8\linewidth}{Reviewer:  #1}  }}
\newcommand{\brscomment}[1]{  \colorbox{blue}{  \parbox{.8\linewidth}{BRS:  #1}  }}
\newcommand{\psrcomment}[1]{  \colorbox{yellow}{  \parbox{.8\linewidth}{PSR:  #1}  }}
\newcommand{\zkcomment}[1]{  \colorbox{cyan}{  \parbox{.8\linewidth}{ZK:  #1}  }}
\newcommand{\ctcomment}[1]{
\colorbox{blue}{  \parbox{.8\linewidth}{CT:  #1}  }}
\newcommand{\swcomment}[1]{ \colorbox{yellow}{ \parbox{.8\linewidth}{ SW: #1}
}}
\newcommand{\shaycomment}[1]{  \colorbox{yellow}{  \parbox{.8\linewidth}{SBC:  #1}  }}
\newcommand{\jlundcomment}[1]{  \colorbox{cyan}{  \parbox{.8\linewidth}{J:  #1}  }}
\newcommand{\kdscomment}[1]{  \colorbox{ceil}{  \parbox{.8\linewidth}{KDS:  #1}  }}
\newcommand{\lkfcomment}[1]{  \colorbox{yellow}{  \parbox{.8\linewidth}{LF:  #1}  }}
\newcommand{\yfcomment}[1]{  \colorbox{brown}{  \parbox{.8\linewidth}{YF:  #1}  }}
\else
\newcommand{\nahocomment}[1]{ }
\newcommand{\shcomment}[1]{ }
\newcommand{\alvincomment}[1]{ }
\newcommand{\jbgcomment}[1]{ }
\newcommand{\ffcomment}[1]{ }
\newcommand{\yhcomment}[1]{ }
\newcommand{\jjmcomment}[1]{ }
\newcommand{\hhecomment}[1]{ }
\newcommand{\tncomment}[1]{ }
\newcommand{\mnicomment}[1]{ }
\newcommand{\prcomment}[1]{ }
\newcommand{\ascomment}[1]{ }
\newcommand{\vecomment}[1]{ }
\newcommand{\halcomment}[1]{ }
\newcommand{\kgcomment}[1]{ }
\newcommand{\brscomment}[1]{ }
\newcommand{\reviewercomment}[1]{ }
\newcommand{\zkcomment}[1]{ }
\newcommand{\jszcomment}[1]{ }
\newcommand{\ctcomment}[1]{ }
\newcommand{\swcomment}[1]{ }
\newcommand{\psrcomment}[1]{ }
\newcommand{\vancomment}[1]{ }
\newcommand{\shaycomment}[1]{ }
\newcommand{\jlundcomment}[1]{ }
\newcommand{\kdscomment}[1]{ }
\newcommand{\lkfcomment}[1]{ }
\newcommand{\mjpcomment}[1]{ }
\newcommand{\fpcomment}[1]{ }
\newcommand{\yfcomment}[1]{ }
\fi

\newcommand{\smallurl}[1]{ \begin{tiny}\url{#1}\end{tiny}}

\usepackage{xcolor}

\definecolor{lightblue}{HTML}{3cc7ea}
\definecolor{CUgold}{HTML}{CFB87C}
\definecolor{grey}{rgb}{0.95,0.95,0.95}
\definecolor{ceil}{rgb}{0.57, 0.63, 0.81}

\usepackage{latexsym}
\usepackage{xcolor,colortbl}
\usepackage{latexsym}
\usepackage{enumitem}
\usepackage{graphicx}
\usepackage{comment}
\usepackage[T1]{fontenc}
\usepackage{diagbox}
\usepackage{amsmath,amssymb,amsthm,amsfonts,enumitem}
\usepackage{dsfont}
\usepackage{lipsum}
\usepackage{multirow}
\usepackage{relsize}
\usepackage{tikz}
\usepackage{pgfplots}
\usepackage{pgfplotstable}
\usepackage{rotating}
\usepackage{adjustbox}
\usepackage{array}
\usepackage{booktabs}
\usepackage{subcaption}
\usepackage{scrextend}
\usepackage{hhline}
\usepackage{csquotes}
\usepackage{sansmath,url}
\usepackage[ruled,vlined,noend,linesnumbered]{algorithm2e}
\usepackage[framemethod=tikz, innerleftmargin=0.3\baselineskip]{mdframed}

\usepackage{url}

\allowdisplaybreaks

\theoremstyle{theoremdd}

\newcommand{\cnpmi}[0]{\textsc{cnpmi}}
\newcommand{\npmi}[0]{\textsc{npmi}}
\newcommand{\inpmi}[0]{\textsc{inpmi}}
\newcommand{\mta}[0]{\textsc{mta}}
\newcommand{\mc}[0]{\textsc{mc}}
\newcommand{\icc}[0]{\textsc{icc}}

\aclfinalcopy 
\def\aclpaperid{475}

\title{Lessons from the Bible on Modern Topics: \\ Low-Resource Multilingual Topic
  Model Evaluation}

\author{Shudong Hao \\
		Computer Science \\
		University of Colorado\\
		Boulder, CO \\
		{\url{shudong@colorado.edu}}
        \And
        Jordan Boyd-Graber \\   
        Computer Science, iSchool, \\ \textsc{lsc}, and \textsc{umiacs},\\
        University of Maryland \\
        College Park, MD \\
        {\url{jbg@umiacs.umd.edu}}
        \And
        Michael J. Paul \\
		Information Science \\
		University of Colorado\\
		Boulder, CO \\
		{\url{mpaul@colorado.edu}}
}

\date{}

\begin{document}
\maketitle
\begin{abstract}
Multilingual topic models enable document analysis across languages through coherent multilingual summaries of the data.  
However, there is no standard and effective metric to evaluate the quality of multilingual topics.
We introduce a new intrinsic evaluation of multilingual topic models that
correlates well with human judgments of multilingual topic coherence as well as performance in downstream applications.
Importantly, we also study evaluation for low-resource languages.
Because standard metrics fail to accurately measure topic quality when robust external resources are unavailable,
we propose an adaptation model that improves the accuracy and reliability of these metrics in low-resource settings.
\end{abstract}

\section{Introduction}
\label{intro}

Topic models provide a high-level view of the main
themes of a document collection~\cite{Boyd-Graber-17}.
Document collections, however, are
often not in a single language, driving the development of
\textbf{multilingual} topic models.  These models discover topics that
are consistent across languages, providing useful tools for
multilingual text analysis~\cite{VulicSTM15}, such as detecting
cultural differences~\cite{GutierrezSLMG16} and bilingual dictionary
extraction~\cite{LiuDM15}.

Monolingual topic models can be evaluated through
likelihood~\cite{WallachMSM09} or coherence~\cite{NewmanLGB10}, but
topic model evaluation is not well understood in multilingual settings.
Our contributions are two-fold.
We introduce an improved intrinsic evaluation metric for multilingual topic models,
called Crosslingual Normalized Pointwise Mutual Information (\cnpmi{}, Section~\ref{sec:eval}).
We explore the behaviors of \cnpmi{} at both the model and topic levels
with six language pairs and varying model specifications.
This metric correlates well with human judgments
and crosslingual classification results
(Sections~\ref{sec:topic-level} and
\ref{sec:model-level}).

We also focus on evaluation in low-resource languages,
which lack large parallel corpora, dictionaries,
and other tools that
are often used in learning and evaluating topic models.
To adapt \cnpmi{} to these settings, we create a coherence estimator
(Section~\ref{sec:coherence-estimator}) that extrapolates statistics
derived from antiquated, specialized texts like the Bible:
often the only resource available for many languages.

\section{Evaluating Multilingual Coherence}
\label{sec:eval}

A multilingual topic contains one topic for each
language.  For a multilingual topic to be meaningful to humans
(Figure~\ref{fig:example}), the meanings should be consistent across
the languages, in addition to coherent within each language
(\textit{i.e.}, all words in a topic are related).

This section describes our approach to evaluating the quality of
multilingual topics.  After defining the multilingual topic model, we
describe topic model evaluation extending standard monolingual
approaches to multilingual settings.

\subsection{Multilingual Topic Modeling}

Probabilistic topic models associate each document in a corpus with a
distribution over latent topics, while each topic is associated with a
distribution over words in the vocabulary.  The most widely used topic
model, latent Dirichlet allocation~\cite[\textsc{lda}]{blei2003}, can
be extended to connect languages.  These extensions require additional
knowledge to link languages together.

One common encoding of multilingual knowledge is \textbf{document
	links} (indicators that documents are parallel or comparable), used
in polylingual topic models~\cite{MimnoWNSM09,NiSHC09}.  In these
models, each document $d$ indexes a tuple of parallel/comparable
language-specific documents, $d^{(\ell)}$, and the language-specific
``views'' of a document share the document-topic distribution $\theta_d$.  The
generative story for the document-links model is: {
	\setlength{\interspacetitleruled}{0pt}
	\setlength{\algotitleheightrule}{0pt}
	\begin{algorithm}
		\small
		\For{each topic $k$ and each language $\ell$}{
			Draw a distribution over words $\phi_{\ell k} \sim \textrm{Dirichlet}(\beta)$\;
		}
		\For{each document tuple $d=\left(d^{(1)},\ldots,d^{(L)}\right)$}{
			Draw a distribution over topics $\theta_d \sim \textrm{Dirichlet}(\alpha)$\;
			\For{each language $\ell=1,\ldots,L$}{
				\For{each token $t\in d^{(\ell)}$}{
					Draw a topic $z_n\sim\theta_d$\;
					Draw a word $w_n\sim\phi_{\ell z}$\;
				}
			}
			
		}
	\end{algorithm}
}

Alternatively, word translations~\cite{JagarlamudiD10}, concept links~\cite{GutierrezSLMG16,YangBR17}, and multi-level
priors~\cite{KrstovskiSK16} can also provide multilingual
knowledges.  Since
the polylingual topic model is the most common approach for building
multilingual topic
models~\cite{VulicSM13,VulicSTM15,LiuDM15,KrstovskiS16}, our study
will focus on this model.

\subsection{Monolingual Evaluation}

Most automatic topic model evaluation metrics use co-occurrence
statistics of word pairs from a reference corpus to evaluate topic
coherence, assuming that coherent topics contain words that often
appear together~\cite{NewmanLGB10}.  The most
successful~\cite{LauNB14} is normalized pointwise
mutual information \cite[\npmi{}]{bouma}.  \npmi{} compares the joint
probability of words appearing together $\Pr(w_i,w_j)$ to their
probability assuming independence $\Pr(w_i)\Pr(w_j)$, normalized by
the joint probability:
\begin{align}
\label{eq:npmi}
\npmi{}(w_i, w_j) = \frac{\log \frac{\Pr(w_i,w_j)}{\Pr(w_i)\Pr(w_j)}}{\log\Pr(w_i,w_j)}.
\end{align}
The word probabilities are calculated from a {\bf reference corpus},
$\mathcal{R}$, typically a large corpus such as Wikipedia that can
provide meaningful co-occurrence patterns that are independent of the
target dataset.

The quality of topic $k$ is the average \npmi{} of
all word pairs $(w_i,w_j)$ in the topic:
\begin{align}
\hspace{-5pt}
\npmi{}_k = \frac{-1}{\binom{C}{2}}\sum_{i \in \mathcal{W}(k, C)} \sum_{j \neq i} \npmi{}(w_i, w_j),
\end{align}
where $\mathcal{W}(k, C)$ are the $C$ most probable words in
the topic-word distribution $\phi_k$ (the number of words is the
topic's \textbf{cardinality}).  Higher 
$\npmi{}_k$ means the topic's top words are more coupled.

\subsection{Existing Multilingual Evaluations}

While automatic evaluation has been well-studied for monolingual topic
models, there are no robust evaluations for multilingual topic models.
We first consider two straightforward metrics that could be used for
multilingual evaluation, both with limitations. We then propose an
extension of \npmi{} that addresses these limitations.

\paragraph{Internal Coherence.}

A simple adaptation of \npmi{} is to
calculate the monolingual \npmi{} score for each language
independently and take the average.  We refer this as internal \npmi{}
(\inpmi{}) as it evaluates coherence \textit{within} a language.
However, this metric does not consider whether the topic is coherent
\textit{across} languages---that is, whether a language-specific word
distribution $\phi_{\ell_1k}$ is related to the corresponding
distribution in another language, $\phi_{\ell_2k}$.

\paragraph{Crosslingual Consistency.}

Another straightforward measurement is Matching Translation
Accuracy~\cite[\mta{}]{Boyd-Graber:Blei-2009}, which counts the number
of word translations in a topic between two languages using a
bilingual dictionary.  This metric can measure whether a topic is
well-aligned across languages \textit{literally}, but cannot capture
non-literal more holistic similarities across languages.

\begin{figure}[t!]
	\centering
	\includegraphics[width=\linewidth]{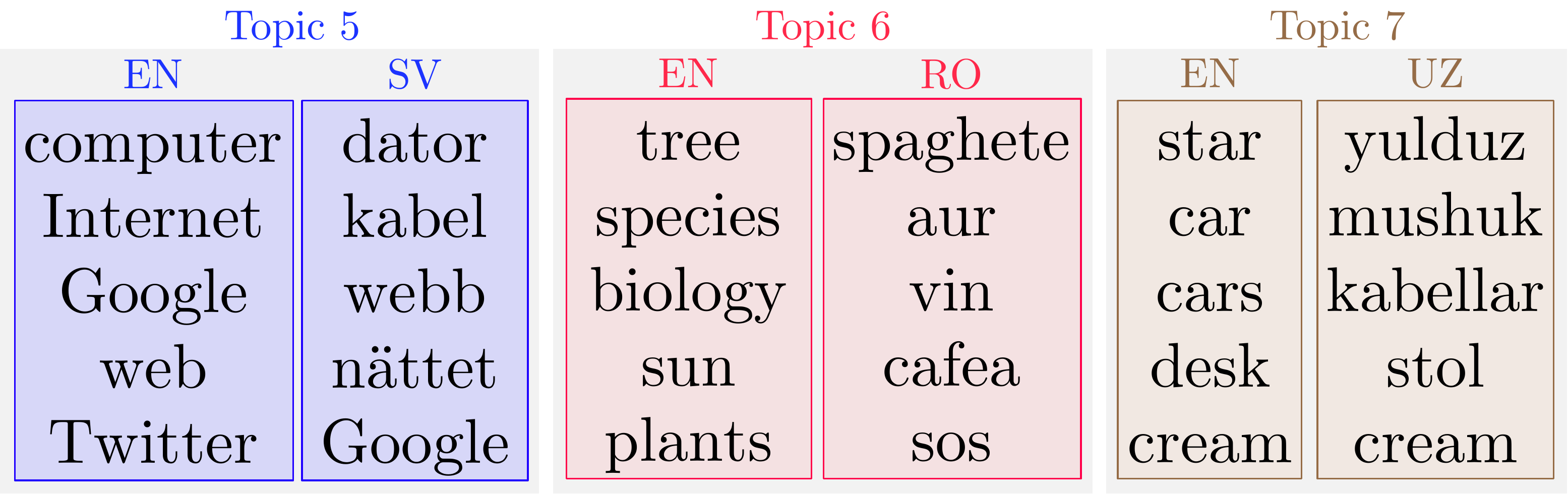}
	\caption{Topic 5 is multilingually coherent: both the English and
		Swedish topics are about \underline{technology}. Topic 6 is about
		\underline{biology} in English but \underline{food} in Romanian, so
		it is low quality although coherent monolingually. Topic 7
		is monolingually incoherent, so it is a low quality topic even if
		it contains word translations.}
	\label{fig:example}
\end{figure}

\subsection{New Metric: Crosslingual \npmi}

We extend \npmi{} to multilingual models, with a metric we
call crosslingual normalized pointwise mutual information (\cnpmi{}).
This metric will be the focus of our experiments.

A multilingually coherent topic means that if $w_{i,\ell_1}$ in language $\ell_1$
and $w_{j,\ell_2}$ in language $\ell_2$ are in the same topic, they should appear
in similar contexts in comparable or parallel corpora
$\mathcal{R}^{(\ell_1,\ell_2)}$.
Our adaptation of \npmi{} is based on the
same principles as the monolingual version,
but focuses on the
co-occurrences of \textit{bilingual} word pairs.
Given a bilingual word pair $\left(w_{i,\ell_1},w_{j,\ell_2}\right)$
the co-occurrence of this word pair is the event
where word $w_{i,\ell_1}$ appears in a document in language~$\ell_1$
and the word $w_{j,\ell_2}$ appears in a comparable or parallel document in language~$\ell_2$.

The co-occurrence probability of each bilingual word pair is:
\begin{align}
\begin{split}
&\Pr\left(w_{i,\ell_1},w_{j,\ell_2}\right) \\ & \triangleq  \frac{\left|\left\lbrace{\mathbf{d}}: w_{i,\ell_1}\in{d}^{(\ell_1)},w_{j,\ell_2}\in{d}^{(\ell_2)}\right\rbrace\right|}{\left|\mathcal{R}^{(\ell_1,\ell_2)}\right|},
\end{split}
\end{align}
where $\mathbf{d}=\left(d^{(\ell_1)}, d^{(\ell_2)}\right)$ is
a pair of parallel/comparable documents in the
reference corpus $\mathcal{R}^{(\ell_1,\ell_2)}$. 
When one or both words in a bilingual pair do not appear in the reference
corpus, the co-occurrence score is zero.

Similar to monolingual settings,
\cnpmi{}
for a bilingual topic $k$ is the average of the
\npmi{} scores of all $C^2$ bilingual word pairs,
\begin{align}\hspace{-8.5pt}
\cnpmi{}(\ell_1,\ell_2,k) = \frac{\sum_{i,j}^{C}\npmi{}\left(w_{i,\ell_1},w_{j,\ell_2}\right)}{C^2}.
\end{align}
It is straightforward to generalize \cnpmi{} from a language pair to multiple
languages by averaging \cnpmi$(\ell_i,\ell_j,k)$ over all language
pairs $(\ell_i,\ell_j)$.

\begin{figure}
	\centering
	\includegraphics[width=\linewidth]{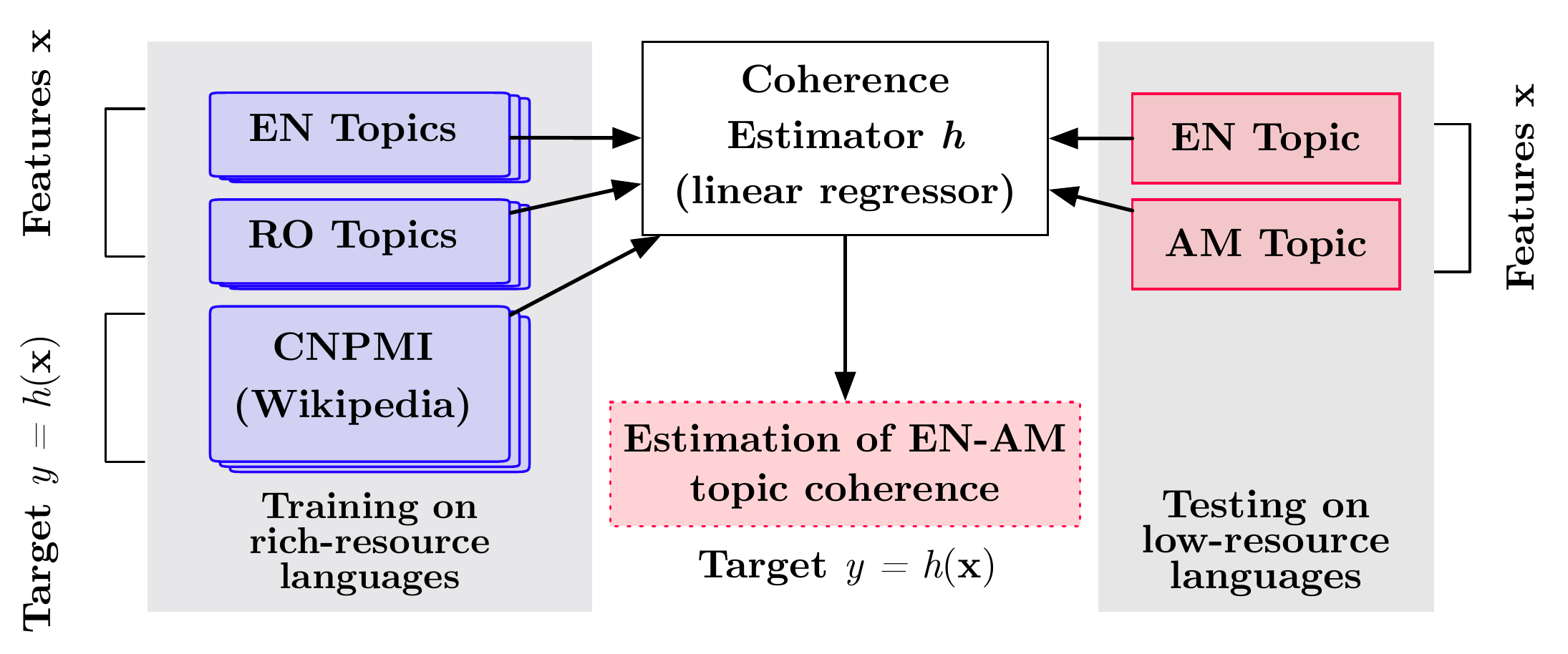}
	\caption{The coherence estimator takes multilingual topics and
		features from them then outputs an estimated topic coherence.}
	\label{fig:calibrator}
\end{figure}

\section{Adapting to Low-Resource Languages}
\label{sec:coherence-estimator}

\cnpmi{} needs a reference corpus for co-occurrence statistics.
Wikipedia, which has good coverage of topics and vocabularies is a
common choice~\cite{JHL16}.  Unfortunately, Wikipedia is often
unavailable or not large enough for low-resource languages. It only
covers $282$
languages,\footnote{\url{https://meta.wikimedia.org/wiki/List_of_Wikipedias}}
and only $249$ languages have more than $1{,}000$ pages: many of pages
are short or unlinked to a high-resource language. Since \cnpmi{}
requires comparable documents, the usable reference corpus is
defined by \emph{paired} documents.

Another option for a parallel reference corpus is the Bible~\cite{resnik-99}, which
is available in most world languages;\footnote{The Bible
	is available in $2{,}530$ languages.} however, it is small
and archaic. It is good at evaluating topics such as
\underline{family} and \underline{religion}, but not ``modern'' topics
like \underline{biology} and \underline{Internet}.  Without reference
co-occurrence statistics relevant to these topics, \cnpmi{} will fail
to judge topic coherence---it must give the ambiguous answer of zero.
Such a score could mean a totally incoherent topic where each word
pair never appears together (Topics~6 in Figure~\ref{fig:example}), or
an unjudgeable topic (Topic~5).

\begin{figure*}[t!]
	\includegraphics[width=\linewidth]{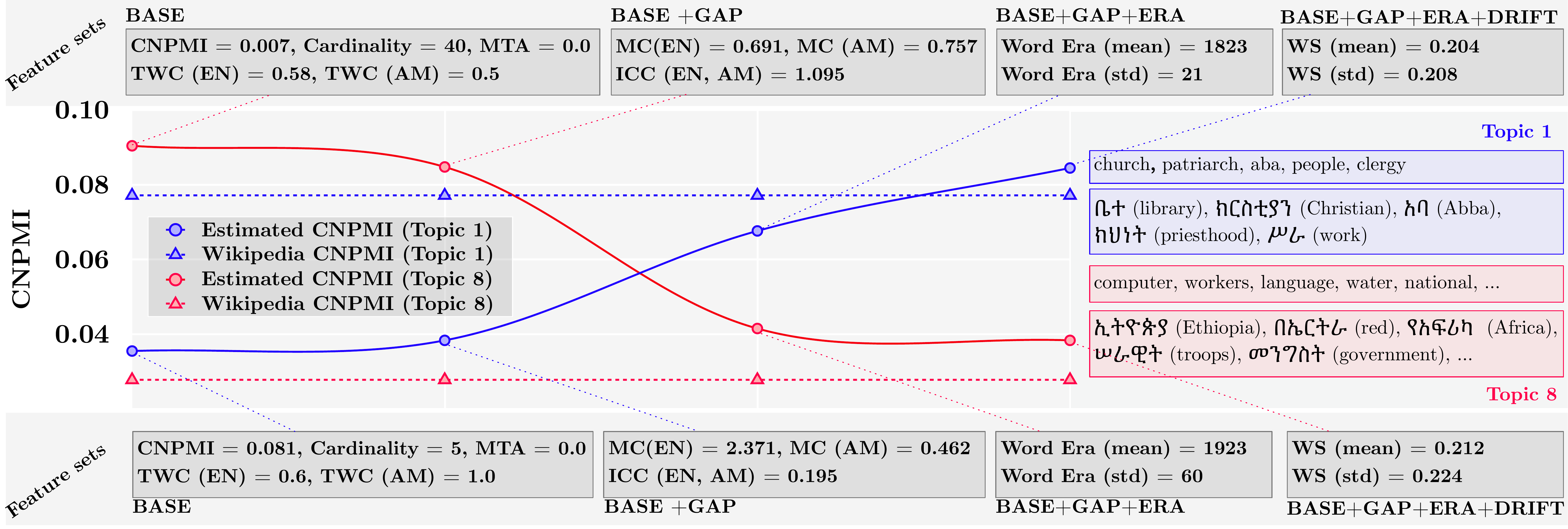}
	\caption{As the estimator adds additional features, the
		estimated topic coherence scores (solid lines) approach to
		Wikipedia \cnpmi{} (dashed lines).}
	\label{fig:example2}
\end{figure*}

Our goal is to obtain a reliable estimation of topic coherence for
low-resource languages when the Bible is the only reference.  We
propose a model that can correct the drawbacks of a Bible-derived
\cnpmi{}.  While we assume bilingual topics paired with
English, our approach can be applied to any high-resource/low-resource
language pair.

We take Wikipedia's \cnpmi{} from high-resource languages as accurate
estimations.  We then build a coherence
\textit{estimator} 
on topics from high-resource languages, with the Wikipedia \cnpmi{} as
the target output.  We use linear regression using the below features.
Given a topic in low-resource language, the estimator produces an
estimated coherence (Figure~\ref{fig:calibrator}).

\subsection{Estimator Features}

The key to the estimator is to find features that capture whether
we should trust the Bible.
For generality, we focus on features independent
of the available resources other than the Bible.
This section describes the features, which we split into four groups.

\paragraph{Base Features (\textsc{base})}
Our base features include information we can collect from
the Bible and the topic model:
cardinality $C$, \cnpmi{} and \inpmi{}, \mta{}, and topic word coverage (\textsc{twc}),
which counts the percentage of topic words in a topic that appear in a reference corpus.

\paragraph{Crosslingual Gap (\textsc{gap})}

A low \cnpmi{} score could indicate a topic pair where each language has
a monolingually coherent topic but that are not about the same theme
(Topic 6 in Figure~\ref{fig:example}). Thus, we add two features to
capture this information using the Bible: mismatch coefficients (\mc{}) and
internal comparison coefficients (\icc{}):
\begin{align}
\mc{}({\ell_1 ; \ell_2,k}) &= \frac{\cnpmi{}(\ell_1,\ell_2,k)}{\inpmi{}(\ell_1,k) + \alpha},\\
\icc{}(\ell_1,\ell_2,k) &= \frac{\inpmi{}(\ell_1,k) + \alpha}{\inpmi{}(\ell_2,k) + \alpha},
\end{align}
where $\alpha$ is a smoothing factor ($\alpha=0.001$ in our
experiments). \mc{} recognizes the gap between crosslingual and monolingual coherence,
so a higher \mc{} score indicates a gap between coherence within and across languages.
Similarly, \icc{} compares monolingual coherence to tell if both languages are coherent:
the closer to $1$ the \icc{} is,
the more comparable internal coherence both languages have.

\paragraph{Word Era (\textsc{era})}

Because the Bible's vocabulary is unable to evaluate modern topics, we
must tell the model what the modern words are.  The \textbf{word era}
features are the earliest usage year
\footnote{\url{https://oxforddictionaries.com/}} for each word in
a topic.  We use both the mean and standard deviation as features.

\paragraph{Meaning Drift (\textsc{drift}).}

The meaning of a word can expand and drift over time.  For example, in the Bible,
``web'' appears in Isaiah~59:5:
\begin{displayquote}
	They hatch cockatrice' eggs, and weave the spider's \textbf{web}.
\end{displayquote}
The word ``web'' could be evaluated correctly in an \underline{animal}
topic. For modern topics, however, Bible fails to capture modern meanings of ``web'',
as in Topic~5 (Figure~\ref{fig:example}).

To address this \textbf{meaning drift}, we use a
method similar to \newcite{HamiltonLJ16}. For each English
word, we calculate the context vector from Bible and from Wikipedia
with a window size of five and calculate the cosine similarity between
them as \textbf{word similarity}.  Similar context vectors mean
that the usage in the Bible is consistent with Wikipedia.
We calculate word similarities for all the English topic words in a
topic and use the average and standard deviation as features.

\subsection{Example}

In Figure~\ref{fig:example2}, Topic~1 is coherent while Topic~8 is
not. From left to right, we incrementally add new feature sets, and
show how the estimated topic coherence scores (dashed lines) approach
the ideal \cnpmi{} (dotted lines). When only using the \textsc{base}
features, the estimator gives a higher prediction to Topic~8 than to
Topic~1. Their low \mta{} and \textsc{twc} prevent accurate
evaluations.  Adding \textsc{gap} does not help much.  However,
$\icc{}(\textsc{en},\textsc{am},k=1)$ is much smaller, which might
indicate a large gap of internal coherence between the two languages.

Adding \textsc{era} makes the estimated scores flip between the two
topics.  Topic~1 has word era of $1823$, much older than Topic 8's
word era of $1923$, indicating that Topic~8 includes modern words the
Bible lacks (\textit{e.g.}, ``computer'').  Using all the features,
the estimator gives more accurate topic coherence evaluations.

\begin{table}
	\centering
	\small
	\setlength\tabcolsep{3pt}
	\begin{tabular}{c|c|ccc}
		\hline
		Pair & Training &\multicolumn{3}{c}{Reference} \\ \cline{3-5}
		&  & \multicolumn{1}{c}{Wikipedia} & The Bible & Wiktionary\\ \hline\hline
		\textsc{en-ro} &  $1{,}272$ & $8{,}126$ & $1{,}189$ & $29{,}836$ \\ \hline
		\textsc{en-sv} & $3{,}378$ & $9{,}067$ & $1{,}189$ & $42{,}953$ \\ \hline
		\textsc{en-am}  & $421$  & $1{,}581$ & $1{,}189$ & $1{,}091$ \\ \hline
		\textsc{en-tl}  & $542$  & $4{,}166$ & $1{,}189$ & $10{,}970$ \\ \hline
		\textsc{en-tr}  & $874$ & $5{,}524$ & $1{,}189$ & $16{,}853$\\ \hline
		\textsc{en-zh}  & $874$ & $10{,}000$ & $1{,}189$ & $22{,}946$\\ \hline
	\end{tabular}
	\caption{Number of document pairs in the training and reference datasets and number of dictionary entries for each language pair.}
	\label{tab:dataset}
\end{table}

\section{Experiments: Bible to Wikipedia}
\label{sec:settings}

We experiment on six languages (Table~\ref{tab:dataset}) from three
corpora: Romanian (\textsc{ro}) and Swedish (\textsc{sv}) from
EuroParl as representative of well-studied and rich-resource
languages~\cite{europarl}; Amharic (\textsc{am}) and Tagalog
(\textsc{tl}) from collected news, as low-resource
languages~\cite{amharicdata,tagalogdata}; and Chinese (\textsc{zh})
and Turkish (\textsc{tr}) from TED Talks 2013~\cite{Tiedemann12},
adding language variety to our experiments. Each language is
paired with English as a bilingual corpus.

Typical preprocessing methods (stemming, stop word removal,
\textit{etc.}) are often unavailable for low-resource languages. For
a meaningful comparison across languages, we do not apply any
stemming or lemmatization strategies, including English, except
removing digit numbers and symbols. However, we remove words that
appear in more than $30\%$ of documents for each language.

Each language pair is separately trained using the
\texttt{MALLET}~\cite{McCallumMALLET} implementation of the
polylingual topic model.
Each experiment runs five Gibbs sampling
chains with $1{,}000$ iterations per chain with twenty topics.
The hyperparameters are set to the default values ($\alpha = 0.1$, $\beta = 0.01$),
and are optimized every $50$ iterations in \texttt{MALLET} using slice
sampling~\cite{wallach-09b}.

\subsection{Evaluating Multilingual Topics}

We use Wikipedia and the Bible as reference corpora for calculating
co-occurrence statistics. Different numbers of Wikipedia articles are
available for each language pair (Table~\ref{tab:dataset}), while the
Bible contains a complete set of $1{,}189$ chapters for all of its
translations~\cite{Christodoulopoulos15}. We use Wiktionary as the
dictionary to calculate \mta{}.

\subsection{Training the Estimator}
\label{sec:est-train}

In addition to experimenting on Wikipedia-based \cnpmi{}, we also
re-evaluate the topics' Bible coherence using our estimator.  In the
following experiments, we use an \mbox{AdaBoost} regressor with linear
regression as the coherence estimator~\cite{friedman2002stochastic,CollinsSS00}.
The estimator
takes a topic and low-quality \cnpmi{} score as input and outputs
(hopefully) an improved \cnpmi{} score.

To make our testing scenario more realistic, we treat one language as
our estimator's test language and train on multilingual topics from
the other languages.  We use three-fold cross-validation over
languages to select the best hyperparameters, including the learning
rate and loss function in \texttt{AdaBoost.R2}~\cite{Drucker97}.

\begin{figure}[t!]
	\centering
	\includegraphics[width=0.48\textwidth]{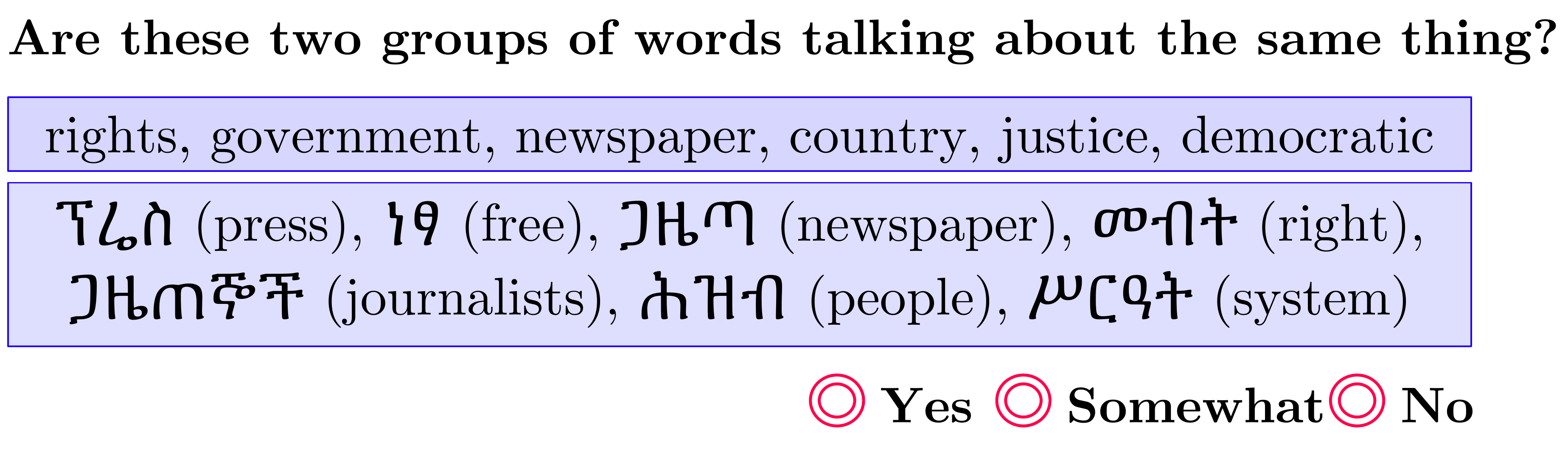}
	\caption{The interface for topic quality judgments. Users read
		the topic first, and make a judgment on whether the words in
		this pair are talking about the same thing. The translations
		are here for illustration; they are not shown to the users.}
	\label{human-inter}
\end{figure}

\section{Topic-Level Evaluation}
\label{sec:topic-level}

We first study \cnpmi{} at the topic level: does a particular topic make
sense?  An effective evaluation should be consistent with
human judgment of the topics~\cite{ChangBGWB09}. In this section, we measure gold-standard
human interpretability of multilingual topics to
establish which automatic measures of topic interpretability work best.

\begin{table}[t!] \centering
	\small
	\setlength\tabcolsep{5pt}
	\begin{tabular}{c|rr|rr|c}
		\hline
		& \multicolumn{2}{c|}{Wikipedia} & \multicolumn{2}{c|}{The Bible} & \multirow{2}{*}{\textsc{mta}} \\ \cline{2-5}
		& \cnpmi{} & \inpmi{} &  \cnpmi{} & \inpmi{} & \\ \hline\hline
		\textsc{en-ro} & $0.490$ & $0.118$ & $-0.096$ & $0.031$ & $\bm{0.592}$ \\
		\textsc{en-sv} & $\bm{0.453}$ & $-0.295$ & $0.164$ & $-0.351$ & $0.248$ \\
		\textsc{en-am} & $0.110$ & $0.019$ & $\bm{0.289}$ & $0.249$ & $0.172$ \\
		\textsc{en-tl} & $\bm{0.512}$ & $0.277$ & $0.166$ & $0.002$ & $0.289$ \\
		\textsc{en-tr} & $0.664$ & $0.243$ & $0.209$ & $-0.246$ & $\bm{0.677}$ \\
		\textsc{en-zh} & $\bm{0.436}$ & $0.297$ & $0.274$ & $0.157$ & $0.411$ \\ \hline
	\end{tabular}
	\caption{Pearson correlations between human judgments and \cnpmi{} are higher than \inpmi{}, while \mta{} correlations are comparable to \cnpmi{}.}
	\label{tab:hj-table}
\end{table}

\subsection{Task Design}

Following monolingual coherence evaluations~\cite{LauNB14}, we present
topic pairs to bilingual CrowdFlower users.  Each task is a topic pair
with the top ten topic words ($C=10$) for each language.  We ask if
both languages' top words in a multilingual topic are talking about
the same concept (Figure~\ref{human-inter}), and make a judgment on a
three-point scale---coherent (2 points), somewhat coherent (1 point),
and incoherent (0 points). To ensure the users have adequate language
competency, we insert several topics that are easily identifiable as
incoherent as a qualification test.

We randomly select sixty topics from each language pair ($360$ topics
total), and each topic is judged by five users. We take the average
of the judgment points and calculate Pearson correlations with
the proposed evaluation metrics (Table~\ref{tab:hj-table}). \npmi{}-based
scores are separately calculated from each reference corpus.

\begin{table}[t!]
	
	\centering
	\small
	\centering
	\begin{tabular}{crccc}
		\hline
		\textbf{Test} & \textbf{Bible} & \multicolumn{3}{c}{\textbf{Train}} \\ \hline\hline
		& & \textsc{ro+sv} & \textsc{zh+tr} & \textsc{ro+sv+zh+tr} \\  \cline{3-5}
		\textsc{am} &  $-0.015$ & $0.332$ & $0.315$ &  $0.333$ \\
		\textsc{tl} &  $-0.309$ & $0.767$ & $0.631$ &  $0.705$ \\ \hline
		& & \textsc{am+tl} & \textsc{zh+tr} & \textsc{am+tl+zh+tr} \\  \cline{3-5}
		\textsc{ro} &  $-0.269$ & $0.736$ & $0.681$ &  $0.713$ \\
		\textsc{sv} &  $0.000$ & $0.787$ & $0.645$ &  $0.683$ \\ \hline
		& & \textsc{ro+sv} & \textsc{am+tl} & \textsc{ro+sv+am+tl} \\  \cline{3-5}
		\textsc{zh} &  $0.217$ & $0.751$ & $0.732$ &  $0.741$ \\
		\textsc{tr} &  $0.113$ & $0.680$ & $0.642$ &  $0.666$ \\ \hline
	\end{tabular}
	\caption{Correlations between the
		Wikipedia-based \cnpmi{} and the Bible-based \cnpmi{},
		before and after using the coherence estimator, at the topic level. Strong
		correlations indicate that the estimator improves \cnpmi{}
		estimates.}
	\label{tab:est-topic}
\end{table}

\subsection{Agreement with Human Judgments}

\cnpmi{} (the extended metric) has higher correlations with human
judgments than \inpmi{} (the naive adaptation of monolingual \npmi{}),
while \mta{} (matching translation accuracy) correlations are
comparable to \cnpmi{}.

Unsurprisingly, when using Wikipedia as the reference, the
correlations are usually higher than when using the Bible. The Bible's
archaic content limits its ability to estimate human judgments in
modern corpora (Section~\ref{sec:coherence-estimator}).

Next, we compare \cnpmi{} to two baselines: \inpmi{} and \mta{}. As
expected, \cnpmi{} outperforms \inpmi{} regardless of reference corpus
overall, because \inpmi{} only considers monolingual coherence. \mta{}
has higher correlations than \cnpmi{} scores from the Bible, because the
Bible fails to give accurate estimates due to limited topic
coverage. \mta{}, on the other hand, only depends on dictionaries,
which are more comprehensive than the Bible. It is also possible that 
users are judging coherence based on
translations across a topic pair, rather than the overall coherence,
which would closely correlate with \mta{}.

\subsection{Re-Estimating Topic-Level Coherence}
\label{sec:estimator}

The Bible---by itself---produces \cnpmi{} values that do not correlate
well with human judgments (Table~\ref{tab:hj-table}).  After
training an estimator (Section~\ref{sec:est-train}), we calculate 
Pearson's correlation between Wikipedia's \cnpmi{} and the estimated
topic coherence score (Table~\ref{tab:est-topic}).  A higher
correlation with Wikipedia's \cnpmi{} means more accurate coherence.

As a baseline, the correlation of Bible-based \cnpmi{} without
adaptation has negative and near-zero correlations with
Wikipedia;\footnote{Normally one would not estimate \cnpmi{} on
	rich-resource languages using low-resource languages. For
	completeness, however, we also include these situations.}  it does
not capture coherence.  After training the estimator, the correlations
become stronger, indicating the estimated scores are closer to
Wikipedia's \cnpmi{}.

\begin{figure}[t]
	\centering
	\includegraphics[width=\linewidth]{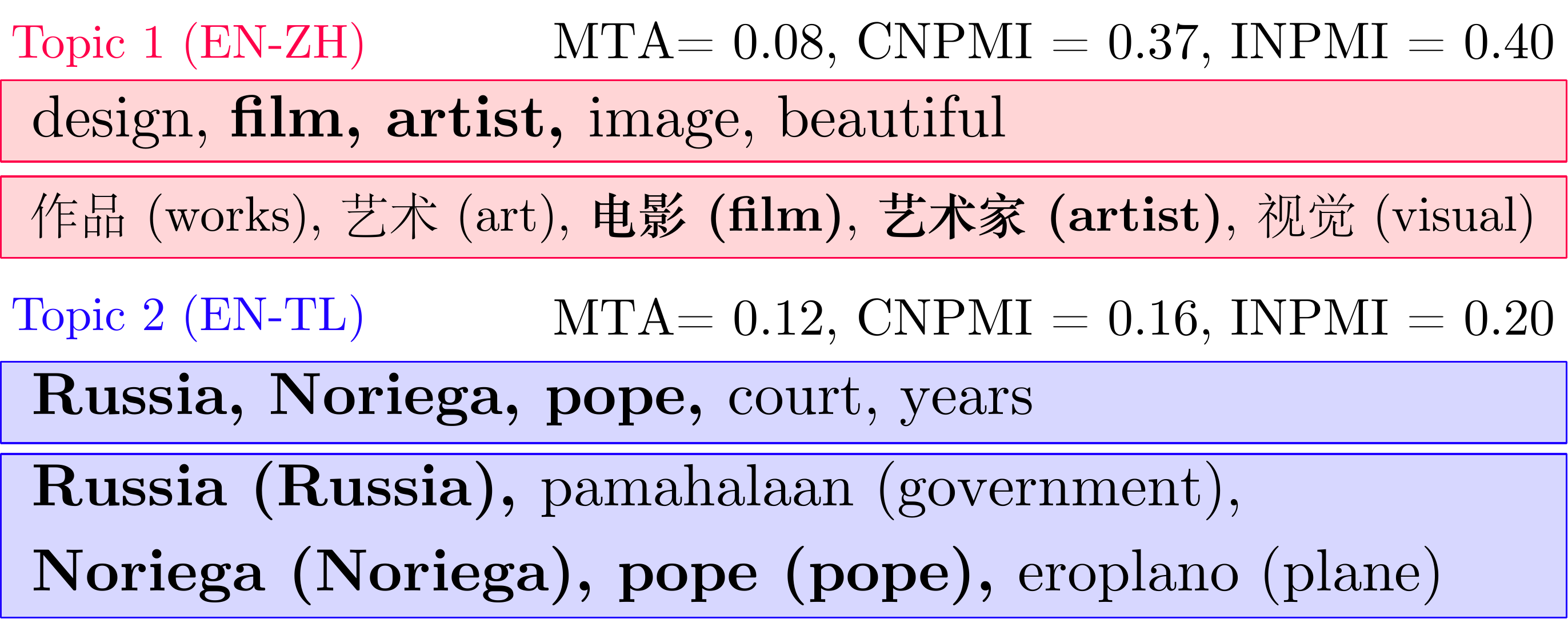}
	\caption{\mta{} fails to capture semantically related words
		(Topic 1) and only looks at translation pairs regardless of
		internal coherence (Topic 2).}
	\label{fig:mtafailexample}
\end{figure}

\subsection{When \mta{} Falls Short}

We analyze \mta{} from two aspects---the inability to
capture semantically-related \textit{non-translation} topic words, and
insensitivity to cardinality---to show why \mta{} is not an ideal
measurement, even though it correlates well with human judgments.

\paragraph{Semantics}
We take two examples with \textsc{en-zh} (Topic 1) and \textsc{en-tl}
(Topic 2) in Figure~\ref{fig:mtafailexample}. Topic~1 has fewer
translation pairs than Topic~2, which leads to a lower \mta{} score
for Topic 1. However, all words in Topic 1 talk about \underline{art},
while it is hard to interpret Topic~2.  Wikipedia 
\cnpmi{} scores reveals Topic~1 is more coherent.  Because
our experiments are on datasets with little divergence between the
themes discussed across languages, this is uncommon for us but could
appear in noisier datasets.

\begin{figure*}[t]
	\centering
	\includegraphics[width=\linewidth]{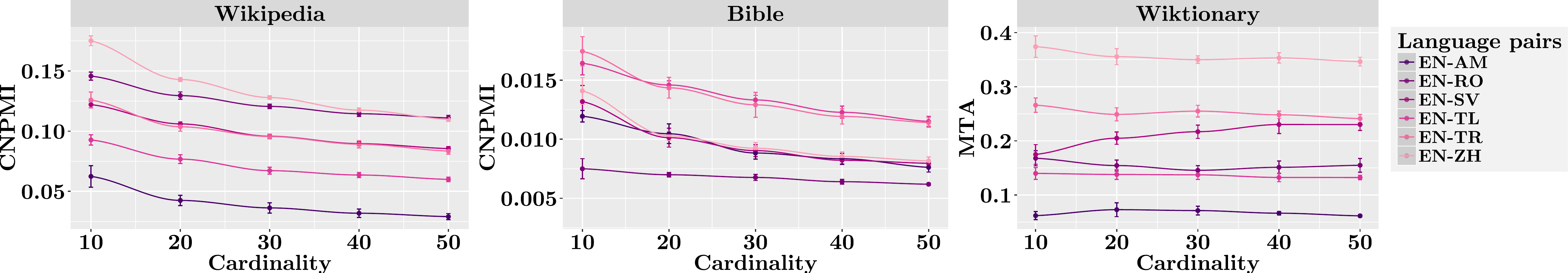}
	\caption{Increasing cardinality of topic pairs makes it harder
		to judge the coherence. Decreasing \cnpmi{} scores reflect
		the diminished interpretability of topics, while \mta{}
		scores do not.}
	\label{fig:add-card}
\end{figure*}

\paragraph{Cardinality}  Increasing cardinality diminishes a topic's coherence~\cite{JHL16}.
We vary the cardinality of topics from ten to fifty at intervals of
ten (Figure~\ref{fig:add-card}).  As cardinality increases, more
low-probability and irrelevant words appear the topic, which lowers
\cnpmi{} scores.  However, \mta{} stays stable or increases with
increasing cardinality.  Thus, \mta{} fails to fulfill a critical
property of topic model evaluation.

Finally, \mta{} requires a comprehensive multilingual dictionary,
which may be unavailable for low-resource languages.
Additionally,
most languages often only have one dictionary,
which makes it problematic to use the same resource (a
language's single multilingual dictionary) for training and evaluating
models that use a dictionary to build multilingual
topics~\cite{HuZEB14}.   Given these concerns, we continue the paper's focus on
\cnpmi{} as a data-driven alternative to \mta{}.  However, for many
applications \mta{} may suffice as a simple, adequate evaluation
metric.

\begin{figure}[t!]
	\centering
	\includegraphics[width=\linewidth]{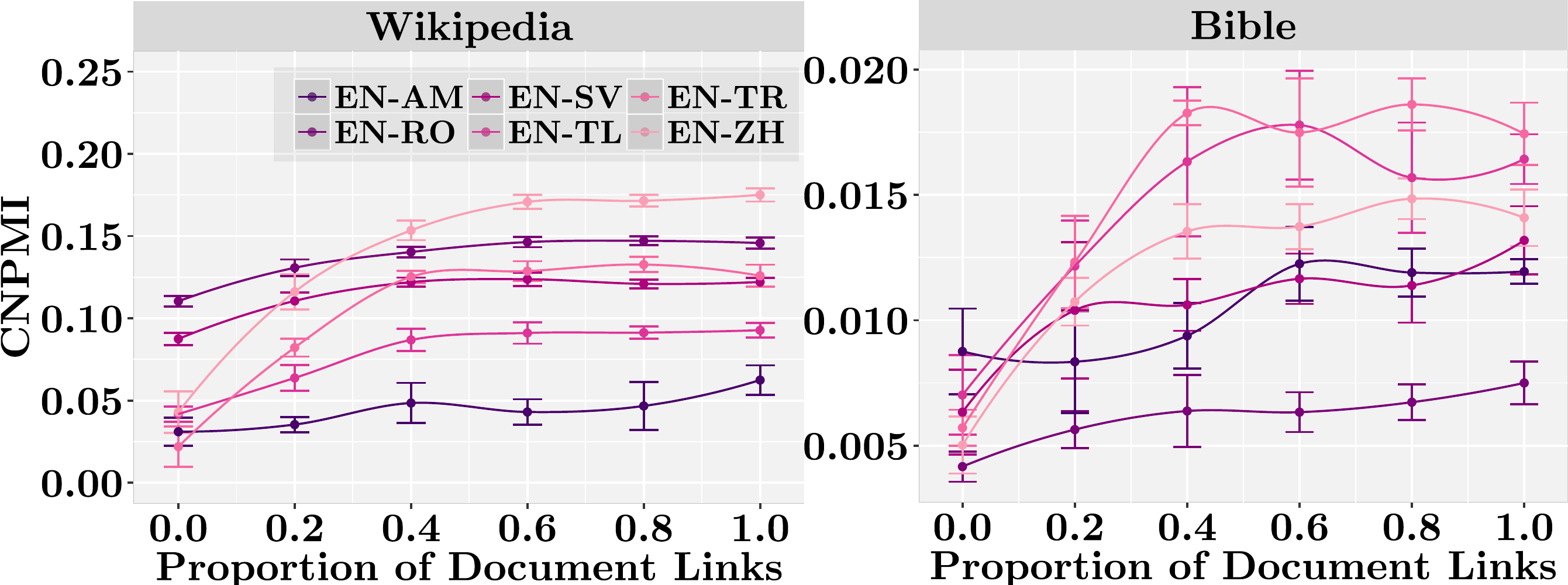}
	\caption{Adding more document links to the model produces more
		multilingually coherent topics. \cnpmi{} captures this
		improvement.}
	\label{fig:add-links}
\end{figure}

\section{Model-Level Evaluation}
\label{sec:model-level}

While the previous section looked at individual topics,  we also care about
how well \cnpmi{} characterizes the quality of \emph{models}
through an average of a model's constituent topics.

\subsection{Training Knowledge}
\label{sec:add-link}

Adding more knowledge to multilingual topic models improves
topics~\cite{HuZEB14}, so an effective evaluation should reflect
this improvement as knowlege is added to the model.
For polylingual topic models, this knowledge takes the form of the
\emph{number} of linked documents.

We start by experimenting with no multilingual knowledge: no document
pairs share a topic distribution $\theta_d$ (but the documents are in
the collection as unlinked documents).  We then increase the number of
document pairs that share $\theta_d$ from $20\%$ of the corpus to
$100\%$.  Fixing the topic cardinality at ten, \cnpmi{} captures the
improvements in models (Figure~\ref{fig:add-links}) through a higher
coherence score.

\begin{table}
	\centering
	\small
	\begin{tabular}{crccc}
		\hline
		\textbf{Test} & \textbf{Bible} & \multicolumn{3}{c}{\textbf{Train}} \\ \hline\hline
		& & \textsc{ro+sv} & \textsc{zh+tr} & \textsc{ro+sv+zh+tr} \\ \cline{3-5}
		\textsc{am} &  \ \ \ $0.607$ & $0.677$ & $0.707$ &  $0.694$ \\
		\textsc{tl} &  $0.796$ & $0.875$ & $0.924$ &  $0.918$ \\ \hline
		& & \textsc{am+tl} & \textsc{zh+tr} & \textsc{am+tl+zh+tr} \\  \cline{3-5}
		\textsc{ro} &  $0.631$ & $0.912$ & $0.919$ &  $0.931$ \\
		\textsc{sv} &  $0.797$ & $0.959$ & $0.848$ &  $0.878$ \\ \hline
		& & \textsc{ro+sv} & \textsc{am+tl} & \textsc{ro+sv+am+tl} \\  \cline{3-5}
		\textsc{zh} &  $0.907$ & $0.918$ & $0.951$ &  $0.939$ \\
		\textsc{tr} &  $0.911$ & $0.862$ & $0.898$ &  $0.887$ \\ \hline
	\end{tabular}
	\caption{At the model level, the estimator improves correlations between \cnpmi{}
		and downstream classification for all languages except for Turkish.}
	\label{tab:est-model}
\end{table}

\subsection{Agreement with Machines}
\label{sec:add-links}

Topic models are often used as a feature extraction technique for downstream machine
learning applications, and topic model evaluations should reflect
whether these features are useful~\cite{RamageHNM09}.  For each model, we apply a
document classifier trained on the model parameters to test whether
\cnpmi{} is consistent with classification accuracy.

\begin{figure}[t!]
	\centering
	\includegraphics[width=\linewidth]{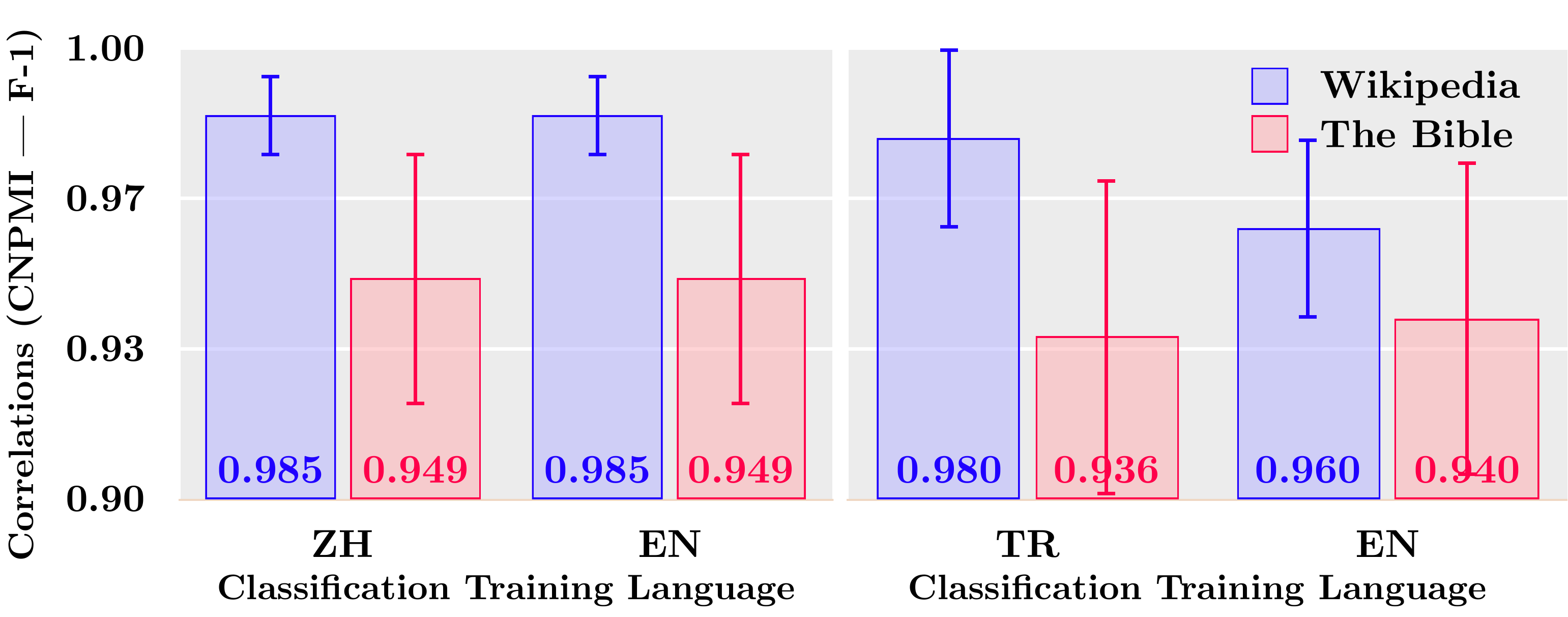}
	\caption{Pearson correlation between classification F1 scores
		and \cnpmi{}: both \cnpmi{} data sources predict whether a
		classifier using topic features will work well, but
		Wikipedia has slightly higher correlation with lower variance.}
	\label{fig:mlc}
\end{figure}

Specifically, we want our classifier to transfer information from
training on one language to testing on another~\cite{SmetTM11,HeymanVM16}.  We train a classifier
on one language's documents, where each document's feature vector is
the document-topic distribution $\theta_d$.  We apply this to
\textsc{ted} Talks, where each document is labeled with multiple
categories. We choose the most frequent seven categories across the
corpus as labels,\footnote{\underline{design}, \underline{global
		issues}, \underline{art}, \underline{science},
	\underline{technology}, \underline{business}, and
	\underline{culture}} and only have labeled documents in one side of
a bilingual topic model.  \cnpmi{} has very strong correlations with
classification results, though using the Bible as the reference corpus
gives slightly lower correlation---with higher variance---than
Wikipedia (Figure~\ref{fig:mlc}).

\subsection{Re-Estimating Model-Level Coherence}

In Section~\ref{sec:estimator}, we improve Bible-based \cnpmi{} scores
for individual topics.  Here, we show the estimator also improves
model-level coherence.  We apply the estimator on the models created
in Section~\ref{sec:add-links} and calculate the correlation between
estimated scores and Wikipedia's \cnpmi{} (Table~\ref{tab:est-model}).

\begin{figure}[t!]
	\centering
	\includegraphics[width=\linewidth]{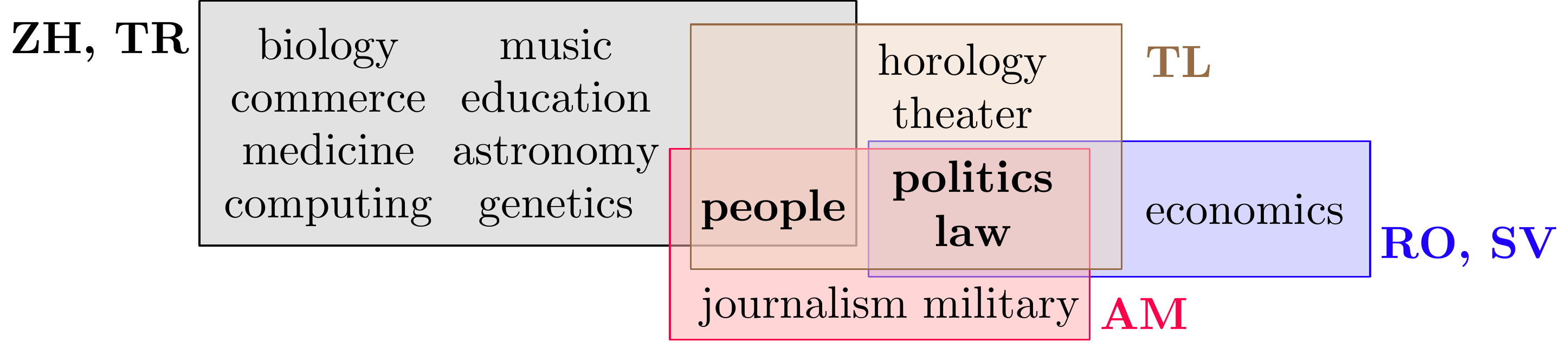}
	\caption{The overlap of topics and domain: only one out of nine
		Turkish and Chinese topics have domain overlap with Tagalog and
		Amharic topics.  This hinders the Turkish estimator from capturing
		model-level properties.}
	\label{fig:overlap}
\end{figure}

The coherence estimator substantially improves scores except for
Turkish: the correlation is better \emph{before} applying the
estimator ($0.911$). We suspect a lack of overlap between topics
between Turkish and languages other than Chinese is to blame
(Figure~\ref{fig:overlap}); the features used by the estimator do not
generalize well to other kinds of features; training on many languages
pairs would hopefully solve this issue.  Turkish is also
morphologically rich, and our preprocessing completely ignores
morphology.

\subsection{Reference Size}
\label{sec:add-ref}

One challenge with low-resource languages is that even if Wikipedia is available,
it may have too few documents to accurately calculate coherence.
As a final analysis, we examine how the reliability of \cnpmi{} degrades
with a smaller reference corpus.

\begin{figure}[t!]
	\centering
	\includegraphics[width=\linewidth]{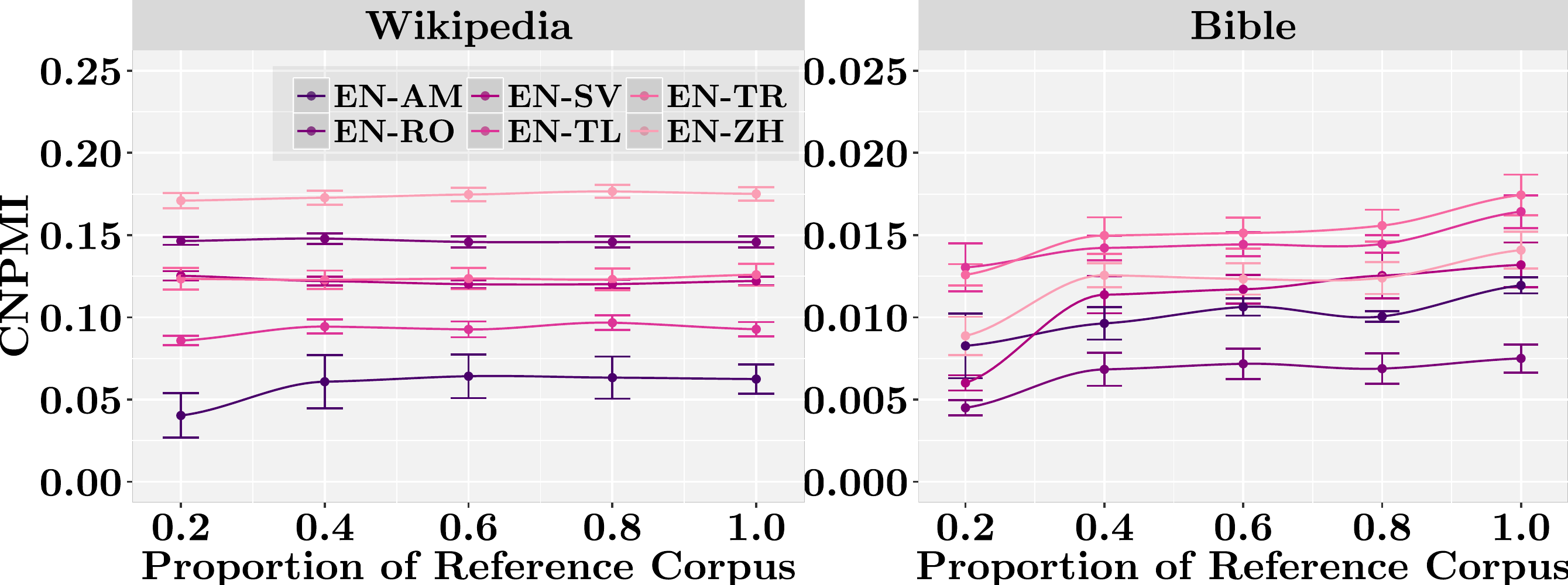}
	\caption{\cnpmi{} is stable once the number of reference
		documents is large enough (around five thousand documents).}
	\label{fig:add-ref}
\end{figure}

We randomly sample
$20\%$ to $100\%$ of document pairs from the reference
corpora and evaluate the polylingual topic model with all document links
(Figure~\ref{fig:add-ref}), again fixing the cardinality as $10$.

\cnpmi{} is stable across different amounts of reference documents,
as long as the number of reference documents is sufficiently large.
If there are too few reference documents (for example, $20\%$ of Amharic Wikipedia is only $316$ documents),
then \cnpmi{} degrades.

\section{Related Work}
\label{sec:related}

\paragraph{Topic Coherence} 
Many coherence metrics based on co-occurrence statistics have been proposed besides \npmi{}.
Similar metrics---such as asymmetrical
word pair metrics~\cite{MimnoWTLM11} and combinations of existing
measurements~\cite{LauNB14,RoderBH15}---correlate well with human
judgments. \npmi{} has been the current gold standard for evaluation
and improvements of monolingual topic
models~\cite{Pecina10,NewmanBB11}.

\paragraph{External Tasks} 
Another approach is to use a model for predictive tasks: the better the results are
on external tasks, the better a topic model is assumed to be.
A common task is held-out likelihood~\cite{WallachMSM09,JagarlamudiD10,FukumasuEX12},
but as \newcite{ChangBGWB09} show, this does not always reflect
human interpretability.
Other specific tasks have also been used,
such as bilingual dictionary extraction~\cite{LiuDM15,MaN17},
cultural difference deteciton~\cite{GutierrezSLMG16},
and crosslingual document clustering~\cite{VulicSTM15}.

\paragraph{Representation Learning}

Topic models are one example of a broad class of techniques of
learning representations of documents~\cite{bengio-13}.  Other
approaches learn respresentations at the
word~\cite{klementiev-12,vyas-15}, paragraph~\cite{mogadala-16}, or
corpus level~\cite{sogaard-15}.  However, neural representation
learning approaches are often data hungry and not adaptable to
low-resource languages.  The approaches here could help improve the
evaluation of all multilingual representation learning
algorithms~\cite{schnabel-15}.

\section{Conclusion}
\label{sec:conclusion}

We have provided a comprehensive analysis of topic model evaluation in multilingual settings,
including for low-resource languages.  While evaluation is an important area of topic model research,
no previous work has studied evaluation of multilingual topic models.
Our work provided two primary contributions to this area, including
a new intrinsic evaluation metric, \cnpmi{},
as well as a model for adapting this metric to low-resource languages without large reference corpora.

As the first study on evaluation for multilingual topic models,
there is still room for improvement and further applications. 
For example, human judgment is more difficult to measure than in monolingual settings,
and it is still an open question on how to design a reliable and accurate survey for multilingual quality judgments. 
As a measurement of multilingual coherence,
we plan to extend \cnpmi{} to high-dimensional representations,
\textit{e.g.,} multilingual word embeddings,
particularly in low-resource languages~\cite{Ruder17}.

\section*{Acknowledgement}

We thank the anonymous reviewers for their insightful and constructive
comments. Hao has been supported under subcontract
to Raytheon \textsc{bbn} Technologies, by \textsc{darpa} award
\textsc{hr0011-15-c-0113}. Boyd-Graber and Paul were supported by
\textsc{nsf} grant \textsc{iis}-1564275. Any opinions, findings, conclusions, or
recommendations expressed here are those of the authors and do not
necessarily reflect the view of the sponsors.

\bibliographystyle{acl_natbib}

\end{document}